# Facial Emotion Detection Using Convolutional Neural Networks and Representational Autoencoder Units

Prudhvi Raj Dachapally

School of Informatics and Computing

Indiana University

***Abstract -*** Emotion being a subjective thing, leveraging knowledge and science behind labeled data and extracting the components that constitute it, has been a challenging problem in the industry for many years. With the evolution of deep learning in computer vision, emotion recognition has become a widely-tackled research problem. In this work, we propose two independent methods for this very task. The first method uses autoencoders to construct a unique representation of each emotion, while the second method is an 8-layer convolutional neural network (CNN). These methods were trained on the posed-emotion dataset (JAFFE), and to test their robustness, both the models were also tested on 100 random images from the Labeled Faces in the Wild (LFW) dataset, which consists of images that are candid than posed. The results show that with more fine-tuning and depth, our CNN model can outperform the state-of-the-art methods for emotion recognition. We also propose some exciting ideas for expanding the concept of representational autoencoders to improve their performance.

## 1. Background and Related Works

The basic idea of using representational autoencoders came from a paper by Hadi Amiri et al. (2016) and they used context-sensitive autoencoders to find similarities between two sentences. Loosely based on that, we expand that idea to the field of vision which will be discussed in the upcoming sections.

There are works that used convolutional neural networks for emotion recognition. Lopes et al. (2015) created a 5-layer CNN which was trained on Cohn – Kanade (CK+) database for classifying six different classes of emotions. A lot of preprocessing steps such spatial and intensity normalization were done before inputting the image to the network for training in this method.

Arushi and Vivek (2016) used a VGG16 pretrained network for this task. Hamester et al. (2015) proposed a 2-channel CNN where the upper channel used convolutional filters, while the lower used Gabor-like filters in the first layer.

Xie and Hu (2017) proposed a different type of CNN structure that used convolutional modules. This module, to reduce redundancy of same features learned, considers mutual information between filters of the same layer, and processes the best set of features for the next layer.

## 2. Methods

### 2.1. Representational Autoencoder Units (RAUs)

We propose two independent methods for the purpose of emotion detection. The first one uses representational autoencoders to construct a unique representation of any given emotion. Autoencoders are a different class of neural networks that can reconstruct their own input in some lower dimensional space. Assume that one image, say of Tom Hanks as in Fig. 1. is sent to this type of network. At first, it generates some random representation in its center-most hidden layer. But if we continue to feed the networks with more and more images of Tom Hanks, the assumption is that the network will be able to develop a unique construct that has the elements of the subject's face encoded in it. Leveraging that intuition, the concept is that an autoencoder network will be able to learn a specific emotion construct for different classes of emotions in the training set.

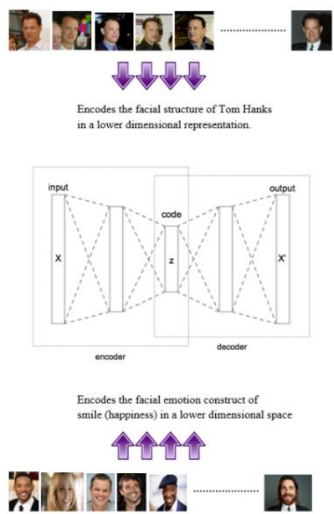

Fig. 1. Intuition behind Representation Autoencoder Units. (Autoencoder image taken from https://en.wikipedia.org/wiki/Autoencoder)

For example, if we feed the autoencoder network 100 different images of people smiling, the network should be able to learn that the feature to encode is the emotional distinctiveness of happiness (smile).

## 2.2. Convolutional Neural Networks

Based on the results of previous publications, we decided to create a CNN on our own and train it from scratch. We created an 8-layer CNN with three convolutional layers, three pooling layers, and two fully connected layers. The structure of the CNN is shown in Fig 2.

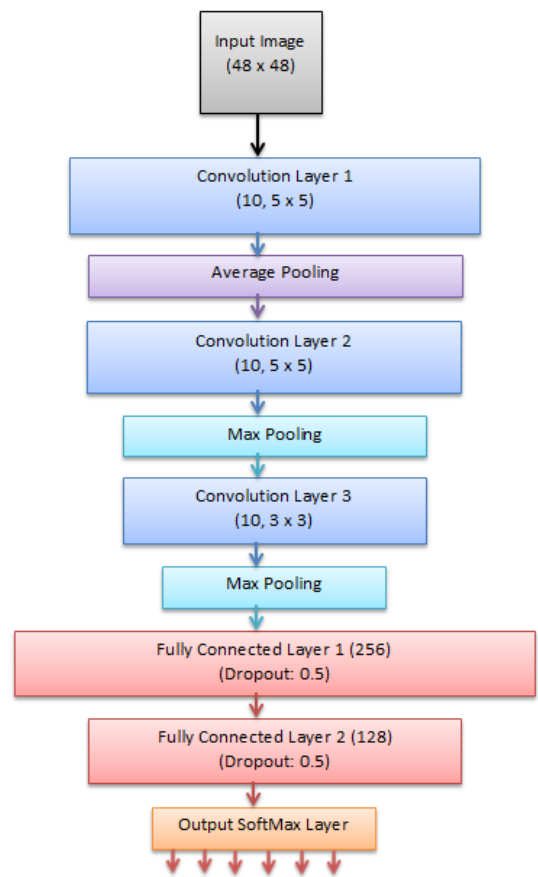

Fig. 2. Proposed Convolutional Neural Network Structure

## 3. Experiments

For RAU method, we developed four different autoencoder networks. The first is a shallow network with only one hidden layer with 300 and 500 nodes, and the second network is layer deeper with another layer of 2800 nodes attached before and after the hidden layer.

We used the Japanese Female Facial Expression (JAFFE) database which has 215 images of 10 different female models posing for 7 emotions. Seventy-five percent of this dataset was used for training, and the rest for testing. The seven emotions in the JAFFE data set are shown in the figure below.

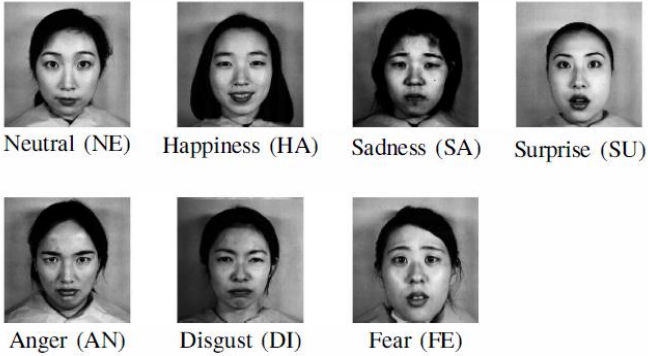

Fig. 3. Seven classes of emotions in the JAFFE dataset (taken from Dennis Hamester et al., 2015)

We resized all the images to 64 x 64 dimensions and first grouped them based on their emotion class. For example, we grouped all the images with anger as their posed emotion and trained the representational units on our autoencoder networks. We did this for all the emotions. Once we had seven distinct representational units for each emotion, for testing, we compared the image with these vectors using a simple cosine distance function.

For the CNN to work well, 159 training images from the JAFFE dataset would not be sufficient. Therefore, we did something along the lines of what AlexNet (Krizhevsky et al., 2012) did for augmenting the training data. Instead of choosing random 48 x 48 patches, we choose all of them. As a result, each 64 x 64 was condensed into 16 – 48 x 48 sized images. Since the original image size was small, more than 95% of all the facial features were conserved in the resultant patches.

Once the augmentation part was done, we had to train the network. All the models were developed using Keras deep learning library. We ran it for 20 epochs at a time, and the best validation accuracy was achieved after 360 iterations. We ran it till 420 iterations, but it was overfitting at that point. Once we had decent validation accuracy, we tested it on the separate test set of 852 images.

Though this was giving decent results, we were not sure how it might work on real-time data (or) candid images. To test the robustness of the model and to perform a cross – database evaluation, 105 images were randomly chosen from the Labeled Faces in the Wild (LFW) dataset, and were sent to 10 different people to gather some subjective ground truth.

## 4. Results

For RAUs, we had 54 test images. The results for this method are given in Table 1. The percentages represent the number of correct predictions by total number of images.

**Representational Autoencoder Units (54 JAFFE test images) Baseline (Random Guessing: 14.28%) ( x – 300/500)(Top 2)**

| Structure | 300 Nodes | 500 Nodes |
|---|---|---|
| 4096 – x – 4096 | 46.29% | 48.19% |
| 4096 – 2800 – x – 2800 – 4096 | 53.70% | 59.25% |

Table 1. Representational Autoencoder Units Model Results for JAFFE test set

The RAU model was also tested on the LFW test set. Those results are shown in Table 2.

**Representational Autoencoder Units (105 LFW test images) Baseline (Random Guessing: 14.28%) (x – 300/500) (Top 2)**

| Structure | 300 Nodes | 500 Nodes |
|---|---|---|
| 4096 – x – 4096 | 41.90% | 44.76% |
| 4096 – 2800 – x – 2800 - 4096 | 50.48% | 48.57% |

Table 2. Representational Autoencoder Units Model Results for LFW test set

Since we are condensing 4096 – dimensional values to 300/500, we took the top two closest distances. This, we think, gave us decent accuracy given the number of examples we trained on. Each emotion had approximately 23 images to train, and we did not use augmented data for this method.

For CNN, we had 852 images from the JAFFE test set and 105 subjectively labeled images from the LFW dataset. Table 3 shows the results for the CNN model.

| Convolutional Neural Network (Trained on 2556 images) Baseline (Random Guessing: 14.28%) | | |
|---|---|---|
| Dataset | Images | Accuracy |
| JAFFE Test Set | 852 | 86.38% |
| LFW (Top 2) | 105 | 67.62% |

Table 3. CNN model results

The CNN model gave good results considering that we only trained 10 filters at each convolution layer. We have some interesting observations from the results (which will be discussed in the next section), and those were one of the main reasons for choosing the top two most probable emotions for the LFW test set.

## 5. Discussion

In this section, we first discuss the results and then some of the observations we found. All the images in the training set were of Japanese women, so, all the samples come from the same ethnicity and of the same gender. Considering that, the results of the autoencoder part were fairly decent. We also combined a number of 64 x 64 images and condensed them into a 300-d (or) 500-d space. On the JAFFE test set, the shallow model predicted 25 correctly out of 54, while the dense model did a little better getting 29 correct. On the LFW data set, this 300 nodes dense autoencoder model predicted 53 correct out of 105, couple of images more than the 500 nodes network.

Since each test example has to be reduced to a 300/500 for testing, we made the number of iterations fixed for each example. The fluctuations in accuracies could be due to the fact that the representations of some images might not have been fully learned. We have a few ideas to improve upon this idea, which will be discussed in the next section.

For the CNN model, the LFW test set predicted 71 out of 105 images correctly. The confusion matrix for these results is shown in Fig 4.

| Confusion Matrix for LFW Test Set (105 Images) | | | | | | | |
|---|---|---|---|---|---|---|---|
| | AN | SA | SU | HA | DI | FE | NE |
| AN (15) | 4 | 2 | 0 | 1 | 3 | 1 | 4 |
| SA (21) | 0 | 20 | 0 | 0 | 1 | 0 | 0 |
| SU (8) | 0 | 1 | 6 | 0 | 0 | 1 | 0 |
| HA (29) | 1 | 3 | 0 | 24 | 1 | 0 | 0 |
| DI (9) | 1 | 0 | 1 | 0 | 7 | 0 | 0 |
| FE (6) | 1 | 0 | 0 | 0 | 1 | 4 | 0 |
| NE (17) | 1 | 4 | 1 | 2 | 0 | 3 | 6 |

Fig. 4. Confusion Matrix for LFW Test Set (CNN)

After gathering all the results, we made a couple of interesting observations about the model predictions and science behind emotion, in general.

Firstly, the boundary between happiness, neutral and sadness is quite thin in the facial structure, contrary to what people might think. This might not be fairly visible in the LFW test set due to limited number of images. But on the JAFFE test set, as seen in Fig 5. , most misclassifications for happiness were neutral and sadness and vice -versa.

| Confusion Matrix for JAFFE Test Set (852 Images) | | | | | | | |
|---|---|---|---|---|---|---|---|
| | AN | SA | SU | HA | DI | FE | NE |
| AN (110) | 102 | 2 | 0 | 0 | 6 | 0 | 0 |
| SA (130) | 2 | 104 | 3 | 8 | 2 | 7 | 4 |
| SU (120) | 0 | 1 | 104 | 0 | 0 | 9 | 6 |
| HA (131) | 2 | 3 | 1 | 110 | 0 | 3 | 12 |
| DI (117) | 8 | 5 | 0 | 2 | 100 | 2 | 0 |
| FE (130) | 1 | 9 | 2 | 1 | 1 | 113 | 3 |
| NE (114) | 0 | 4 | 2 | 5 | 0 | 0 | 103 |

Fig. 5. Confusion Matrix for JAFFE Test Set (CNN)

One of the main reasons for choosing the LFW data set for testing is the candidness of the images. These, as said earlier, are not posed to enact a certain expression. Therefore, most subtleties were leveraged by the network to respond with some decent predictions. Fig 6. shows couple of example predictions that illustrate this border.

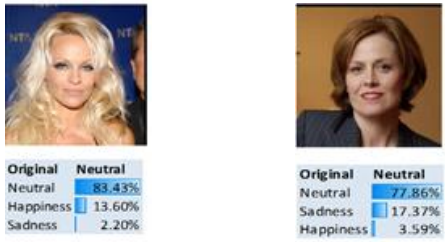

Fig. 6. Predictions illustrating the line between Happiness, Neutral, and Sadness (Percentages show the confidence of the prediction)

In the above figure, consider the second image. The ground truth label for that image was neutral. But on the quick glimpse, we can make an educational guess that the person is actually smiling, so the prediction emotion should be happy. But in general, the emotion is neutral, since one can observe that the person is trying to maintain a "poker" face. The thing we learned about emotions is their perception is subjective to the context created by a third-person's mind.

Another thing about the CNN model we learned is that when an image has a negative emotion, all the top predictions tend to be negative emotions (sadness, fear, disgust, anger). If the given image does not have position vibe (or) emotion to it, its top predicted emotions tend to be negative. Fig. 7 shows a couple of these examples.

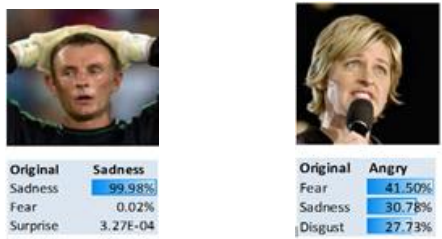

Fig. 7. Predictions illustrating negative predictions (Percentages show the confidence of the prediction)

In the first image, the person is clearly not happy. At the same time, an aglow of anger is not seen in his face. One can estimate that the person in that image is quite disappointed, and that emotion is close to sadness. The next prediction was fear, which was not a positive emotion. In the second image, the correct emotion was hard to guess for the network. But the ground truth was angry. That emotion was not in the top three, but the top predictions of the network were all negative emotions.

Coming back to CNNs, the basic intuition is that the different layers of the neural network learn features that are unique for a specific emotion. Since the filters do not reveal much information when shown as themselves, we applied those filters on an image to observe what kind of features the network learns. Figure 8 shows some of these images, layer by layer. We manually increased the brightness values to make structures visible.

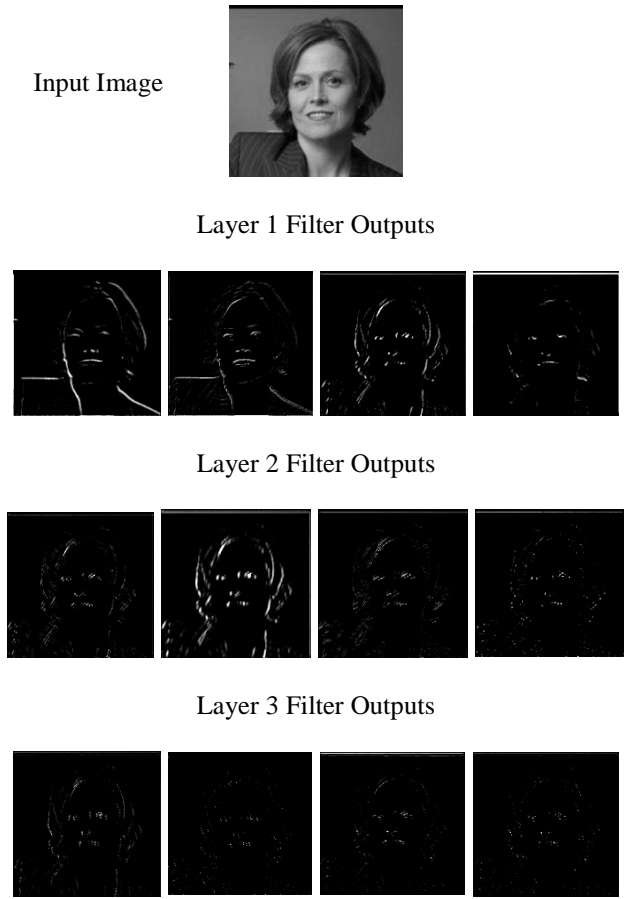

Fig. 8. CNN Filter Outputs

## 6. Conclusion and Future Work

In this work, we introduced two different methods for solving the problem of emotion detection. The first one using autoencoders, though the intuition was right and the idea was fairly original, it did not generate the best results. One of the main reasons we think is since we concatenate all the pixels vertically to feed into the autoencoder, we might be losing some structural integrity of the image. In the future, thank you for the valuable input from Dr. Sven Bambach, we want to replace the normal hidden units with convolution filters in the encoder part, and deconvolution units for the decoder. The second method was using a convolutional neural network, and with three convolutional layers, three pooling layers and two fully – connected layers, we achieved a good accuracy on the JAFFE test set as well as on the LFW test set. We want to work more on the visualizing the learned filters in depth, and in the future, we also want to take a semi-supervised approach by using the predictions made for the LFW images, to train the network with more data, more filters, and more depth.

## 7. Acknowledgements

We would like to thank Professor David Crandall for his valuable input and guidance throughout the entire project. We would like to thank the developers of all the deep learning frameworks for making the works of students, professionals, and researches a little bit easier. We specially thank Dr. Sven Bambach for his valuable input on expanding the idea of representational autoencoder units.

## 8. References

[1] Hadi Amiri et al. "Learning Text Pair Similarity with Context – sensitive Autoencoders", *ACL* 2016.

[2] Andre Teixeira Lopes et al. "A Facial Expression Recognition System Using Convolutional Networks", Vol. 00, pg. 273 – 280, 2015.

[3] Arushi Raghuvanshi and Vivek Choksi, "Facial Expression Recognition with Convolutional Neural Networks", *CS231n Course Projects*, Winter 2016.

[4] Dennis Hamester et al., "Face Expression Recognition with a 2-Channel Convolutional Neural Network", *International Joint Conference on Neural Networks (IJCNN)*, 2015.

[5] Siyue Xie and Haifeng Hu, "Facial expression recognition with FRR – CNN", *Electronic Letters,* 2017, Vol. 53 (4), pg. 235 – 237.

[6] Nima Mousavi et al. "Understanding how deep neural networks learn face expressions", *International Joint Conference on Neural Networks*, July 2016.

[7] Razavian et al. "CNN Features off-the-shelf: an Astounding Baseline for Recognition", *arXiv: 1403.6382v3[cs.CV],* May 2014.

[8] Michael J. Lyons et al . "Coding Facial Expressions with Gabor Wavelets", *3rd IEEE International Conference on Automatic Face and Gesture Recognition*, pp. 200-205 (1998).

[9] Gary B. Huang et al. "Labeled Faces in the Wild: A Database for Studying Face Recognition in Unconstrained Environments." *University of Massachusetts, Amherst, Technical Report 07-49*, October, 2007.

[10] Francois Chollet, Keras, GitHub, *https://github.com/fchollet/keras,* 2015.

[11] Alex Krizhevsky et al. "ImageNet Classification with Deep Convolutional Neural Networks", *Neural Information Processing Systems (NIPS),* 2012.